\documentclass[conference]{IEEEtran}

\usepackage{graphicx}
\usepackage{booktabs}
\usepackage{multirow}
\usepackage{amsmath, amssymb}
\usepackage{siunitx}
\usepackage{enumitem}
\usepackage{url}
\usepackage{xcolor}
\usepackage{subcaption}

\newcommand{\dataset}{PlotChain}

\begin{document}

\title{\dataset: Deterministic Checkpointed Evaluation of Multimodal LLMs on Engineering Plot Reading}

\author{
\IEEEauthorblockN{Mayank Ravishankara}
\IEEEauthorblockA{
Independent Researcher, San Francisco, CA, USA \\
Email: mravisha@alumni.cmu.edu (alt: mayankgowda@gmail.com)
}
}

\maketitle

\begin{abstract}
We present \textbf{PlotChain}, a deterministic, generator-based benchmark for evaluating multimodal large language models (MLLMs) on \emph{engineering plot reading}---recovering quantitative values from classic plots (e.g., Bode/FFT, step response, stress--strain, pump curves) rather than OCR-only extraction or free-form captioning. PlotChain contains \textbf{15 plot families} with \textbf{450 rendered plots} (30 per family), where every item is produced from known parameters and paired with \textbf{exact ground truth} computed directly from the generating process. A central contribution is \emph{checkpoint-based diagnostic evaluation}: in addition to final targets, each item includes intermediate ``cp\_*'' fields that isolate sub-skills (e.g., reading cutoff frequency or peak magnitude) and enable failure localization within a plot family. We evaluate four state-of-the-art MLLMs under a standardized, deterministic protocol (temperature $=0$ and a strict JSON-only numeric output schema) and score predictions using per-field tolerances designed to reflect human plot-reading precision. Under the \texttt{plotread} tolerance policy, the top models achieve \textbf{80.42\%} (Gemini 2.5 Pro), \textbf{79.84\%} (GPT-4.1), and \textbf{78.21\%} (Claude Sonnet 4.5) overall field-level pass rates, while GPT-4o trails at \textbf{61.59\%}. Despite strong performance on many families, frequency-domain tasks remain brittle: bandpass response stays low (\(\leq\)23\%), and FFT spectrum remains challenging. We release the generator, dataset, raw model outputs, scoring code, and manifests with checksums to support fully reproducible runs and retrospective rescoring under alternative tolerance policies.
\end{abstract}


% ============================
\section{Introduction}
% ============================
\label{sec:introduction}

Engineering practice is mediated by plots: Bode magnitude/phase curves, FFT spectra, step responses,
stress--strain diagrams, IV curves, pump characteristic curves, and calibration charts are routinely
used to validate designs, troubleshoot systems, and communicate quantitative results. Progress is
difficult to measure without benchmarks that (i) target \emph{engineering} plot families (not generic
infographic charts), (ii) provide \emph{exact} and auditable ground truth, and (iii) support
\emph{diagnostic} evaluation that distinguishes visual read errors from downstream arithmetic or
reasoning errors.

Existing work on chart and figure understanding has made important advances, but typical benchmarks are not designed to isolate the specific failure modes that arise in engineering plot reading. Early datasets such as FigureQA provide synthetic scientific-style plots with template questions and auxiliary annotations \cite{kahou2017figureqa}. DVQA focuses on bar charts and highlights brittleness to appearance variation and the challenge of answers unique to each chart \cite{kafle2018dvqa}. PlotQA targets scientific plots at scale and emphasizes reasoning and out-of-vocabulary numeric answers \cite{methani2020plotqa}. ChartQA introduces human-written questions that require visual and logical reasoning over charts, often leveraging the underlying chart data table \cite{masry2022chartqa}. More recent MLLM-focused suites (e.g., ChartBench) broaden chart types and question styles for chart comprehension evaluation \cite{xu2023chartbench}, while general multimodal reasoning benchmarks (e.g., MathVista and MMMU) include charts among many heterogeneous modalities and tasks \cite{lu2024mathvista,yue2024mmmu}. These resources are valuable, yet they commonly (a) emphasize generic chart QA or captioning rather than canonical \emph{engineering} plot families, (b) evaluate final answers without intermediate sub-skill probes, and/or (c) rely on data tables or human annotations in ways that make ``ground truth'' less directly attributable to the plotted rendering process \cite{methani2020plotqa,masry2022chartqa,xu2023chartbench}.

In this paper, we introduce \textbf{PlotChain}, a deterministic, generator-based benchmark tailored to \emph{engineering plot reading}. PlotChain renders plot images from known parameters and computes ground truth deterministically from those same parameters, enabling exact verification and eliminating annotation noise. The benchmark spans \textbf{15 plot families} with \textbf{450 items} (30 per family), with controlled difficulty settings (clean/moderate/edge) that remain human-readable but stress common plot-reading challenges such as sparse tick marks, reduced grid cues, tighter axis windows, and ambiguous intersections. Each item pairs a natural-language question with a required \emph{strict JSON} numeric response schema to support automated scoring and reproducible evaluation.

A central novelty is \emph{checkpoint-based diagnostic evaluation}. In addition to final target fields, items may include intermediate ``cp\_*'' fields that correspond to localized plot reads (e.g., reading a cutoff frequency, peak magnitude, intercept, slope region, or axis scale) that are scored separately from final fields. This design turns plot reading into a sequence of verifiable sub-reads, enabling fine-grained attribution: models may succeed at reading intermediate quantities yet fail on derived quantities (or vice versa), which is not observable from a single final-answer score. To reflect realistic human plot interpretation, we score numeric predictions using per-field tolerances (absolute and/or relative), rather than exact string match, and we report both headline pass rates and per-family diagnostic breakdowns.

Finally, reproducibility is treated as a first-class objective. Motivated by broader calls for transparent, reproducible ML evaluation \cite{pineau2021reproducibility}, and benchmark practices that release prompts and raw outputs for auditability \cite{liang2022helm}, PlotChain is distributed with the generator, dataset JSONL, evaluation scripts, raw model outputs, and manifests with checksums and run metadata. This enables exact re-runs when possible and retrospective rescoring under alternative tolerance or parsing policies without re-querying model APIs.

\paragraph{Contributions.}
\begin{itemize}[leftmargin=*,noitemsep,topsep=2pt]
  \item \textbf{Deterministic engineering-plot benchmark:} a generator-based dataset spanning 15 canonical engineering plot families with exact ground truth derived from generation parameters.
  \item \textbf{Checkpoint-based diagnostics:} intermediate ``cp\_*'' fields that isolate plot-reading sub-skills and enable failure localization beyond final-answer evaluation.
  \item \textbf{Standardized, deterministic evaluation protocol:} uniform prompting and decoding controls (temperature $=0$) with strict JSON numeric outputs, coupled with tolerance-based scoring aligned to human plot reading.
  \item \textbf{Reviewer-auditable artifact release:} code, data, raw outputs, and manifests with checksums to support reproducible runs and post-hoc analysis.
\end{itemize}

% ============================
\section{Related Work}
% ============================
\label{sec:related}

Interpreting engineering plots requires extracting quantitative values (often absent from the rendered pixels), mapping visual encodings to axes/units, and performing light numerical reasoning under perceptual uncertainty. Prior work spans (i) chart/plot question answering benchmarks, (ii) chart \emph{derendering} and plot-to-table conversion, (iii) chart summarization/captioning for accessibility, and (iv) broader multimodal benchmarks and evaluation practices.

\subsection{Chart and Plot Question Answering Benchmarks}
Early work studied chart question answering alongside explanation generation, typically assuming access to structured chart specifications or focusing on smaller curated sets \cite{kim2020charts}. Synthetic benchmarks such as FigureQA \cite{kahou2017figureqa} and DVQA \cite{kafle2018dvqa} scaled supervision by programmatically generating charts and questions; these enabled controlled evaluation but often emphasize discrete labels or restricted answer forms and chart styles.

PlotQA \cite{methani2020plotqa} targets scientific plots with real-valued answers and larger variability, highlighting challenges beyond fixed vocabularies and prompting hybrid pipelines that combine perception with structured reasoning. ChartQA \cite{masry2022chartqa} extends to real-world charts (with human-written questions) and emphasizes both visual and logical reasoning, often relying on explicit table extraction or auxiliary parsing of chart elements. Collectively, these datasets demonstrate that chart understanding is neither pure OCR nor generic captioning, but a compositional perception--reasoning problem with brittle failure modes.

\subsection{Derendering and Plot-to-Table Conversion}
A complementary line of work treats charts as a structured graphics modality to be \emph{derendered} into underlying data tables or symbolic representations. Earlier systems and interactive tools addressed chart data extraction as a core primitive for downstream reasoning and accessibility \cite{jung2017chartsense, siegel2016figureseer}. ChartOCR \cite{luo2021chartocr} combines learned and rule-based components and introduces large-scale annotated data for chart elements.

More recent approaches emphasize end-to-end or unified modeling. ChartReader \cite{cheng2023chartreader} proposes a unified framework that supports both derendering and comprehension without heavy manual rule design. Pix2Struct \cite{lee2023pix2struct} pretrains on screenshot-to-structured-output tasks and is frequently used as a strong baseline for chart/plot parsing. UniChart \cite{masry2023unichart} introduces chart-specific pretraining objectives to improve both low-level extraction and high-level reasoning.

DePlot \cite{liu2023deplot} popularizes a modality-conversion strategy: translate plots into a table-like representation and then perform reasoning on the structured output, while also discussing inconsistencies in evaluation metrics across chart types. Benchmarking studies on LVLM chart capability and correction strategies further document systematic chart-specific weaknesses and evaluation pitfalls \cite{huang2024lvlms}.

\subsection{Chart Summarization and Accessibility}
Chart-to-text and summarization benchmarks focus on generating natural-language descriptions of charts, motivated by accessibility and visualization literacy \cite{obeid2020chart2text, kantharaj2022charttotext}. While these tasks overlap with plot understanding, they emphasize narrative faithfulness and content selection rather than numeric extraction accuracy under realistic plot-reading tolerances.

\subsection{Chart-Focused Evaluation in the LVLM Era}
With large vision-language models, several chart-focused evaluations and benchmarks have emerged to probe chart reasoning and robustness \cite{xu2023chartbench, xia2024chartx}. These efforts reinforce the need for evaluation designs that separate (a) visual extraction from (b) reasoning, and that characterize failure modes beyond a single final-answer metric.

Broader multimodal benchmarks, including MMBench \cite{liu2024mmbench}, MathVista \cite{lu2024mathvista}, and MMMU \cite{yue2024mmmu}, evaluate general multimodal understanding and reasoning, but are not designed specifically for quantitative engineering-style plot reading or for checkpoint-based diagnostic analysis. \cite{11317986}.

\subsection{Benchmarking Methodology and Reproducibility}
Standardized evaluation protocols and artifact release practices are increasingly emphasized in the language and multimodal model literature \cite{liang2022helm, pineau2021reproducibility}. Our work aligns with this direction by centering deterministic generation, checksumed artifacts, fixed decoding settings, and storing raw outputs to enable rescoring under revised tolerance policies. We use standard scientific Python tooling commonly cited in reproducible ML pipelines \cite{virtanen2020scipy, harris2020numpy, hunter2007matplotlib}.

% ============================
% \section{Dataset: \dataset{}}
% ============================
% =========================================================
\section{Benchmark Design and Dataset}
\label{sec:benchmark}

\subsection{Task Definition}
\label{sec:task_def}
PlotChain evaluates \emph{engineering plot reading} by multimodal LLMs: given an image of a canonical engineering plot and a natural-language question, the model must return a \emph{single JSON object} whose values are numeric (or \texttt{null} when explicitly instructed).
Unlike OCR-only settings, PlotChain requires interpreting axes (including log scales), reading off curve values, and computing derived quantities consistent with standard engineering practice (e.g., cutoff frequency at $-3$ dB, settling time by a tolerance band, bandwidth between $-3$ dB points).

Each item specifies two categories of target fields:
\begin{itemize}
  \item \textbf{Final fields} (headline): the main numeric answers (e.g., \texttt{cutoff\_hz}, \texttt{settling\_time\_s}).
  \item \textbf{Checkpoint fields} (diagnostics): intermediate reads prefixed \texttt{cp\_} (e.g., \texttt{cp\_mag\_at\_fc\_db}, \texttt{cp\_peak\_value}) that localize failure modes and enable stepwise capability analysis.
\end{itemize}

\subsection{Plot Families}
\label{sec:families}
PlotChain contains 15 plot families (30 items per family; 450 total items). Table~\ref{tab:families} summarizes families, domains, and representative outputs. Families are chosen to reflect common analysis patterns in controls, signals, circuits, mechanical systems, and materials.

\begin{table}[t]
\centering
\caption{PlotChain plot families and representative target variables (final fields; checkpoint fields are prefixed \texttt{cp\_} and omitted for brevity).}
\label{tab:families}
\footnotesize
\begin{tabular}{p{0.28\columnwidth} p{0.22\columnwidth} p{0.34\columnwidth}}
\hline
\textbf{Family} & \textbf{Domain / Axes} & \textbf{Representative final outputs} \\
\hline
\texttt{step\_response} & Controls; time vs response & \texttt{percent\_overshoot}, \texttt{settling\_time\_s}, \texttt{steady\_state} \\
\texttt{bode\_magnitude} & Circuits/controls; log-$f$ vs dB & \texttt{dc\_gain\_db}, \texttt{cutoff\_hz} \\
\texttt{bode\_phase} & Circuits/controls; log-$f$ vs deg & \texttt{cutoff\_hz}, \texttt{phase\_deg\_at\_fq} \\
\texttt{bandpass\_ response} & Filters; log-$f$ vs dB & \texttt{resonance\_hz}, \texttt{bandwidth\_hz} \\
\texttt{time\_waveform} & Signals; time vs voltage & \texttt{frequency\_hz}, \texttt{vpp\_v} \\
\texttt{fft\_spectrum} & Signals; $f$ vs magnitude & \texttt{dominant\_frequency\_ hz}, \texttt{secondary\_frequency\_ hz} \\
\texttt{spectrogram} & Signals; time--freq heatmap & \texttt{f1\_hz}, \texttt{f2\_hz}, \texttt{switch\_time\_s} \\
\texttt{iv\_resistor} & Circuits; $V$--$I$ linear & \texttt{resistance\_ohm} \\
\texttt{iv\_diode} & Circuits; $V$--$I$ exponential & \texttt{target\_current\_a}, \texttt{turn\_on\_voltage\_v \_at\_target\_i} \\
\texttt{transfer\_ characteristic} & Nonlinear blocks; $V_{in}$--$V_{out}$ & \texttt{small\_signal\_gain}, \texttt{saturation\_v} \\
\texttt{pole\_zero} & Systems; complex plane & \texttt{pole\_real}, \texttt{pole\_imag}, \texttt{zero\_real}, \texttt{zero\_imag} \\
\texttt{stress\_strain} & Materials; strain vs stress & \texttt{yield\_strength\_mpa}, \texttt{uts\_mpa}, \texttt{fracture\_strain} \\
\texttt{torque\_speed} & Motors; speed vs torque & \texttt{stall\_torque\_nm}, \texttt{no\_load\_speed\_rpm} \\
\texttt{pump\_curve} & Fluids; flow vs head & \texttt{head\_at\_qop\_m}, \texttt{q\_at\_half\_head\_m3h} \\
\texttt{sn\_curve} & Fatigue; log--log cycles vs stress & \texttt{stress\_at\_1e5\_mpa}, \texttt{endurance\_limit\_mpa} \\
\hline
\end{tabular}
\end{table}

\subsection{Deterministic Generation and Ground Truth}
\label{sec:determinism}
All items are generated by a \emph{frozen} generator that deterministically maps \texttt{(master\_seed, family, index)} to: (i) plot parameters, (ii) a rendered PNG plot image, and (iii) a ground-truth dictionary computed directly from the generating parameters using explicit analytic / numeric baselines (not OCR).
This design yields \emph{verifiable gold labels}: for every numeric target, the paper can (and does) release the corresponding parameterization and baseline used to compute the value.

To maintain human-readability and avoid pathological targets (e.g., irrational or visually implausible fractions), PlotChain \emph{quantizes} ground-truth values to family/field-specific decimal precision (e.g., integer Hz cutoffs, 0.1 dB gains, and task-appropriate rounding for times/ratios). This makes the evaluation reflect realistic plot reading rather than ultra-fine numerical reconstruction.

\subsection{Difficulty and Edge-Case Design}
\label{sec:difficulty}
Within each family, items follow an intended difficulty mixture of approximately \textbf{40\% clean / 30\% moderate / 30\% edge}.
\textbf{Clean} items include standard visual aids (ticks, gridlines, clear axis windows).
\textbf{Moderate} items apply controlled perturbations (e.g., mild noise) while remaining readable.
\textbf{Edge} items remove or weaken visual aids (e.g., tighter axis windows or reduced cues) but are constrained by a manual readability protocol to remain solvable by a human using approximate plot-reading.

\subsection{Dataset Format and Released Artifacts}
\label{sec:format}
PlotChain is distributed as:
\begin{itemize}
  \item a JSONL file (\texttt{plotchain.jsonl}) containing one object per item;
  \item per-family image directories (\texttt{images/<family>/...png});
  \item validation CSVs produced by the generator (row-level and summary-level sanity checks).
\end{itemize}

Each JSONL item includes:
\begin{itemize}
  \item \texttt{id}, \texttt{type}, \texttt{image\_path}, \texttt{question};
  \item \texttt{ground\_truth}: dictionary with both final fields and optional \texttt{cp\_} checkpoint fields;
  \item \texttt{plot\_params}: generating parameters required to reproduce the plot and recompute labels;
  \item \texttt{generation}: metadata including deterministic seed, difficulty tag, and explicit lists of final vs checkpoint fields.
\end{itemize}

\begin{figure*}[t]
\centering
\setlength{\tabcolsep}{2pt}
\renewcommand{\arraystretch}{1.0}

\newcommand{\mont}[1]{\includegraphics[width=0.19\textwidth]{figures/montage/#1}}

\begin{tabular}{ccccc}
\mont{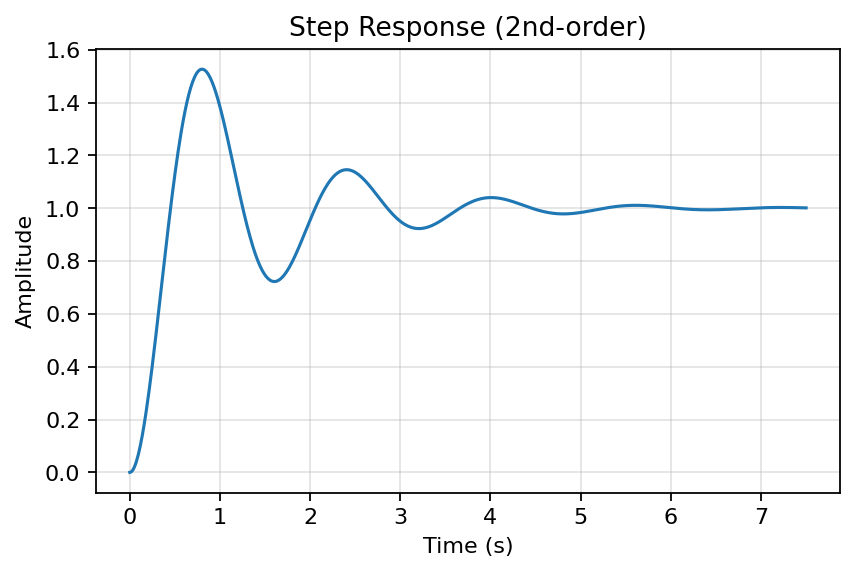} &
\mont{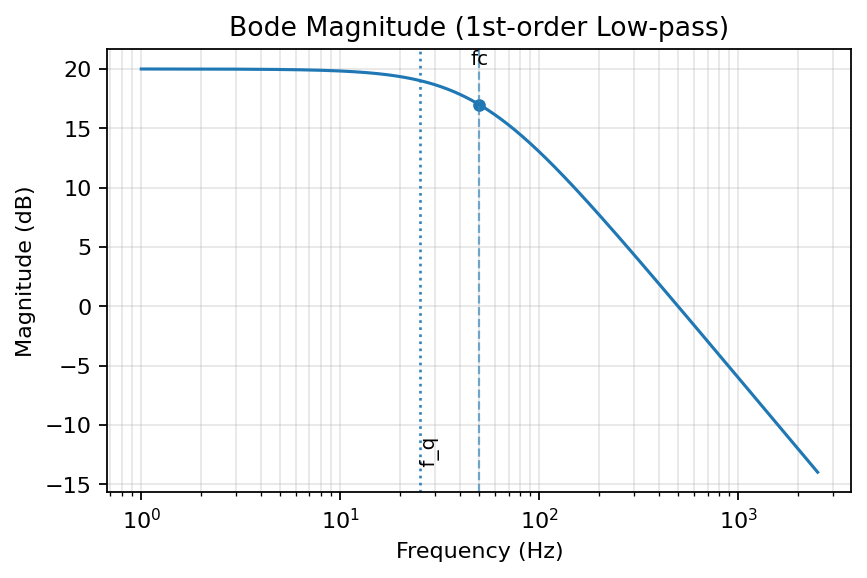} &
\mont{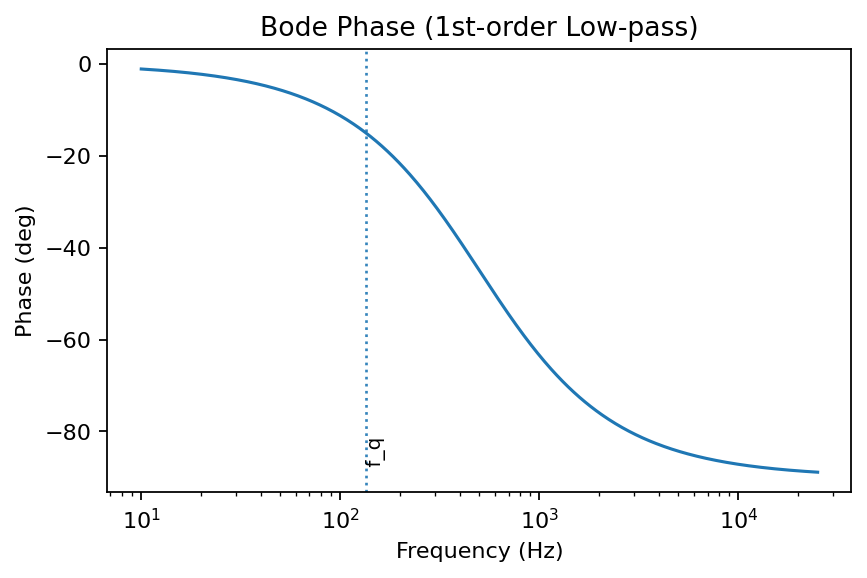} &
\mont{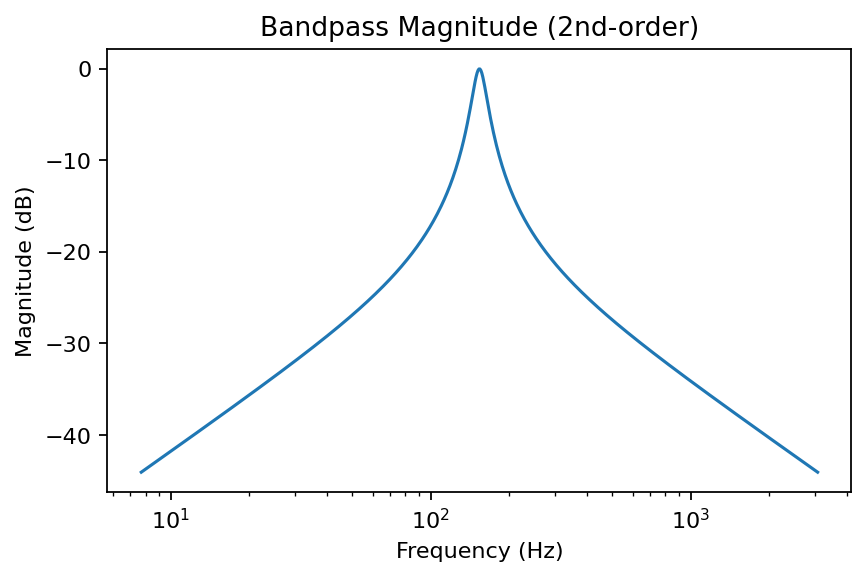} &
\mont{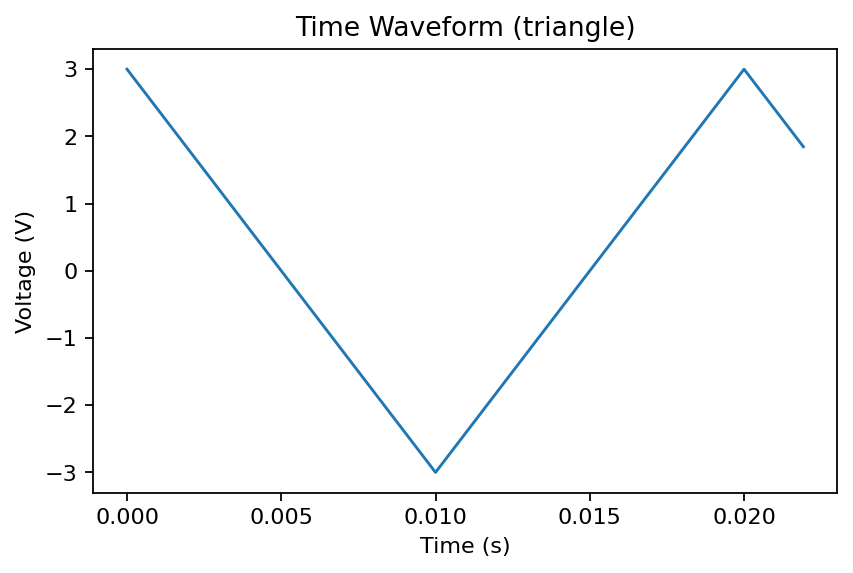} \\

\mont{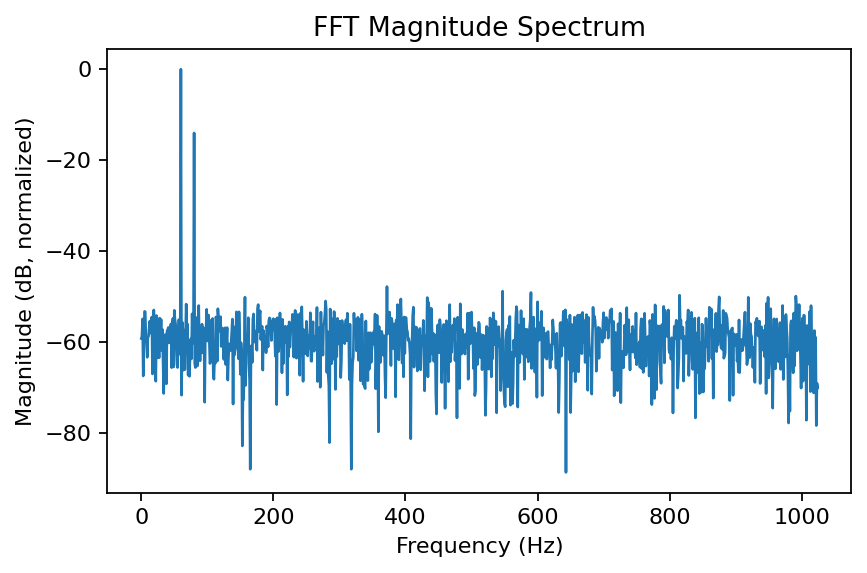} &
\mont{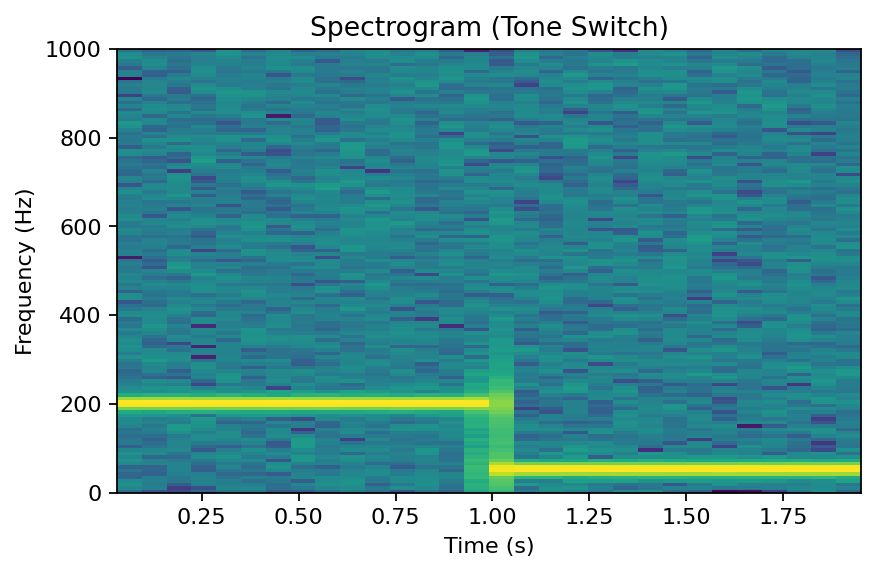} &
\mont{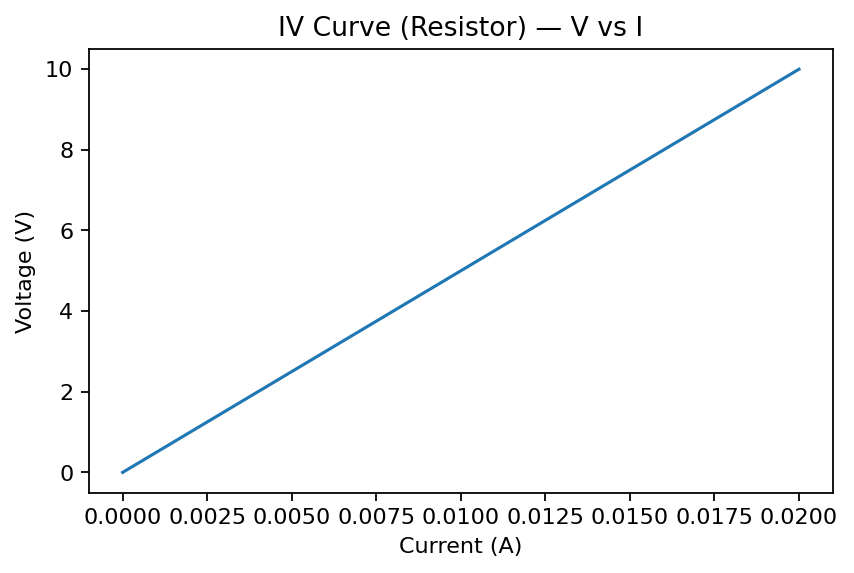} &
\mont{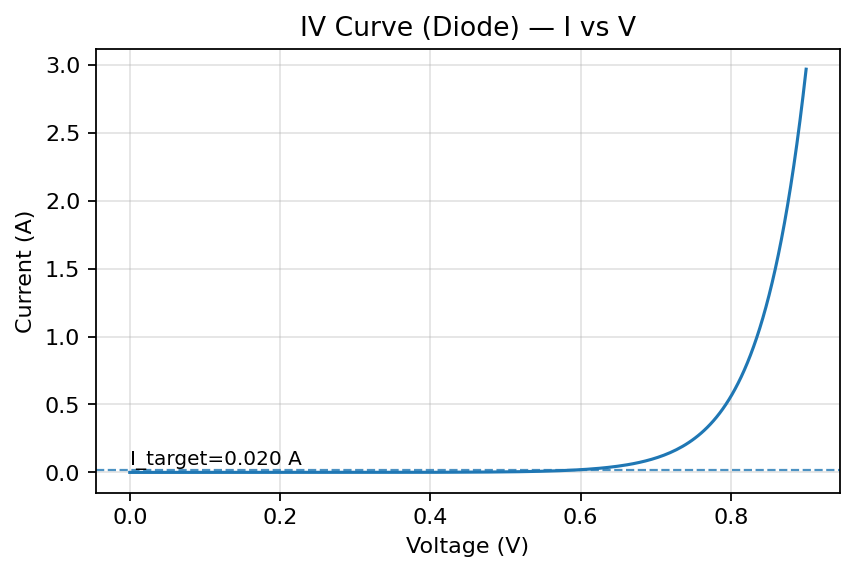} &
\mont{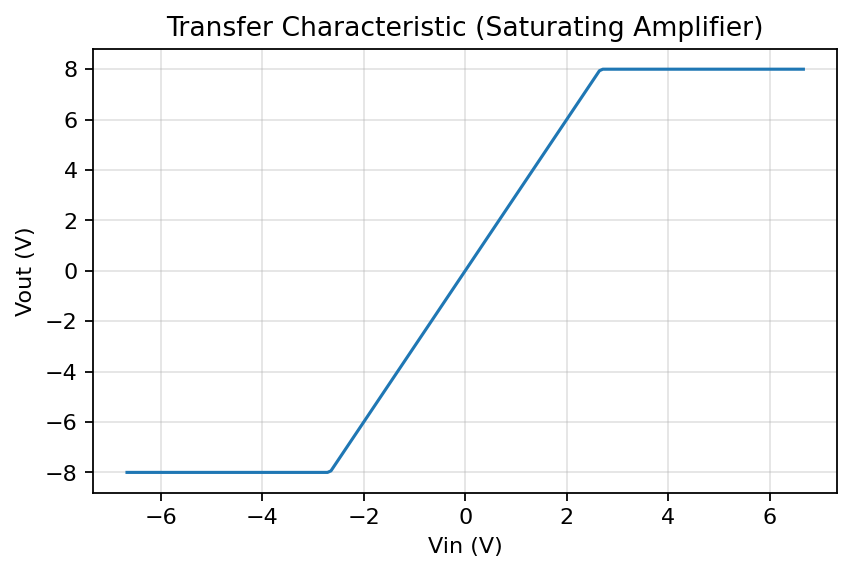} \\

\mont{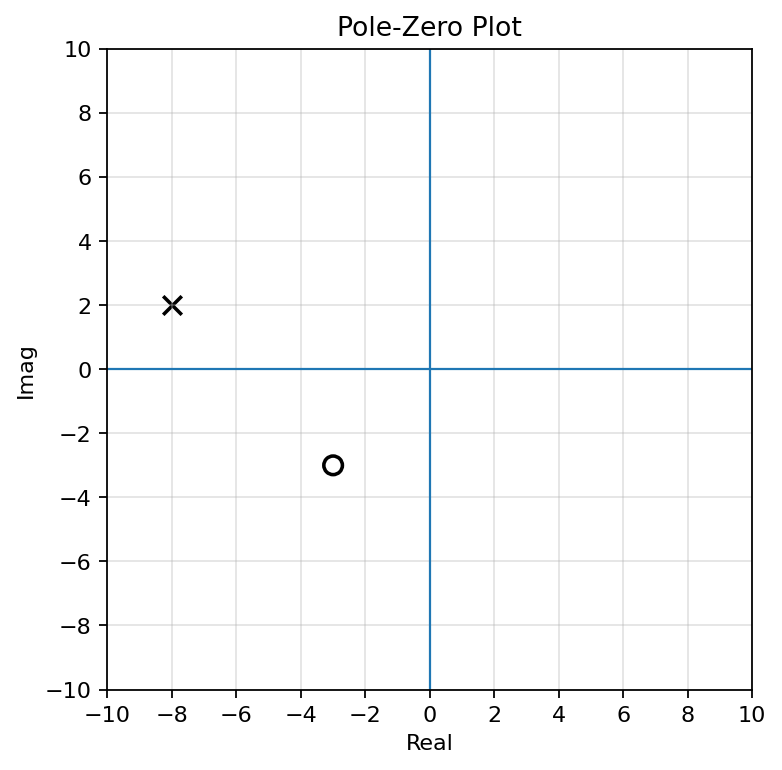} &
\mont{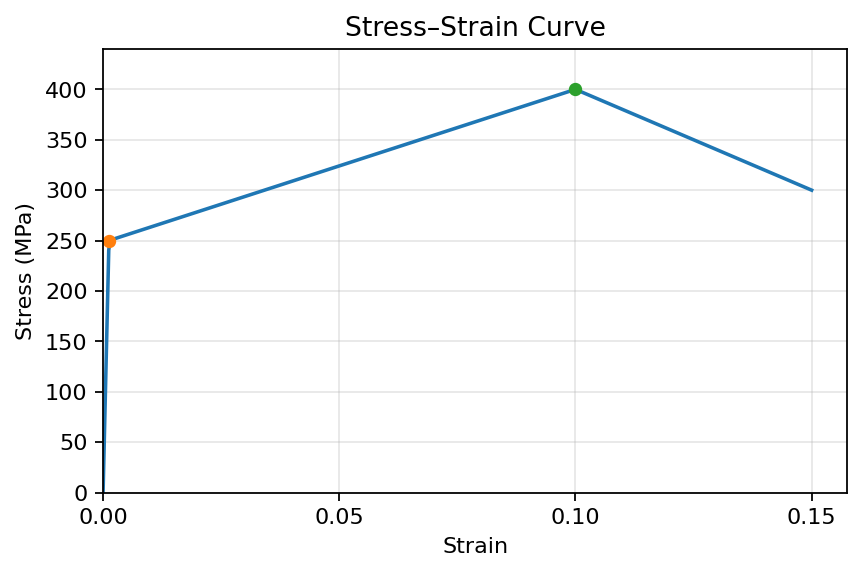} &
\mont{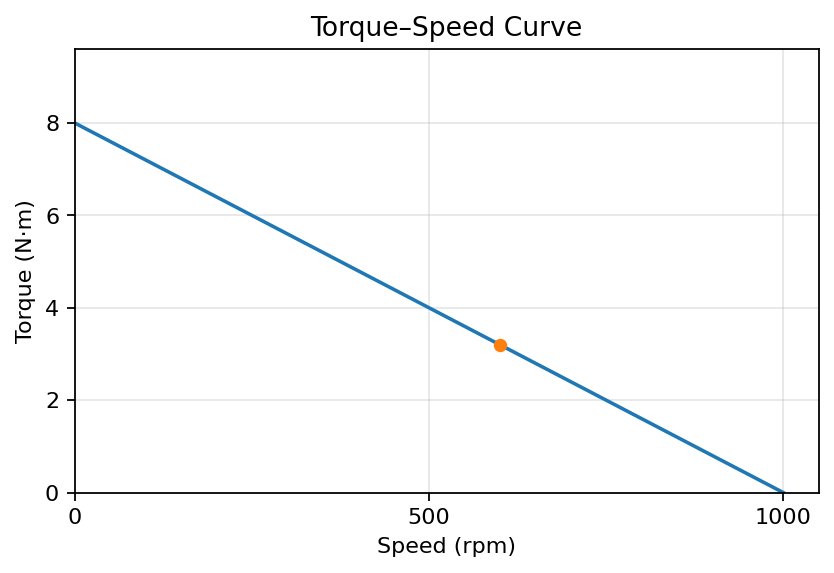} &
\mont{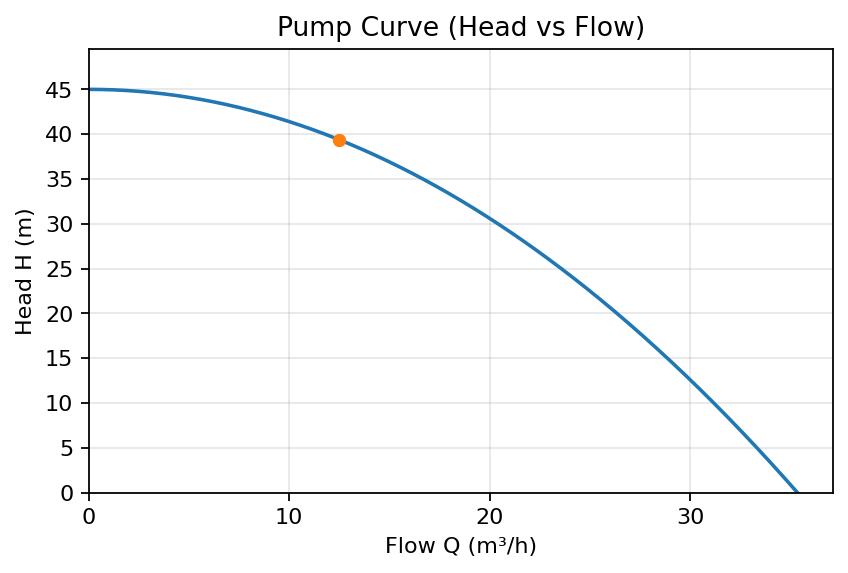} &
\mont{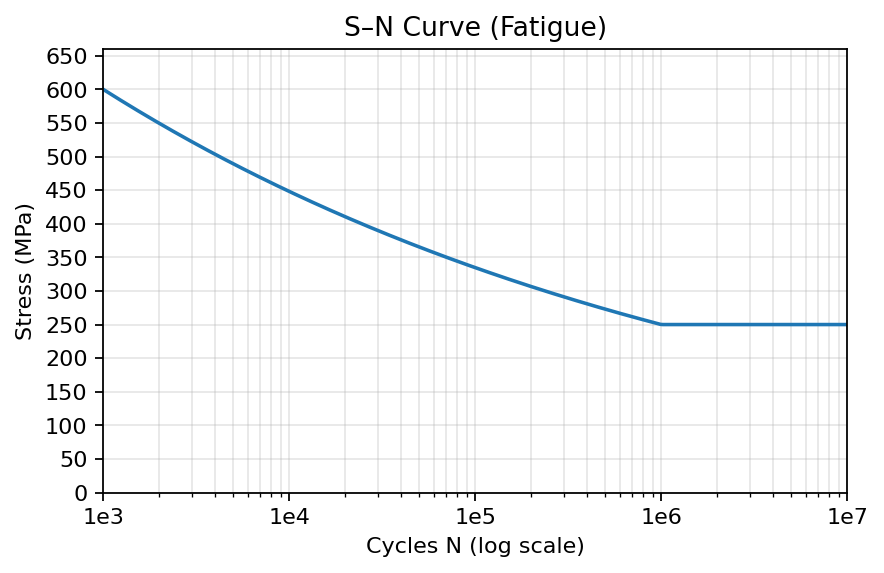} \\
\end{tabular}

\caption{Representative PlotChain samples (one per family). Row 1: Step response; Bode magnitude; Bode phase; Bandpass frequency response; Time-domain waveform. Row 2: FFT magnitude spectrum; Spectrogram; Resistor I--V curve; Diode I--V curve; Transfer characteristic. Row 3: Pole--zero plot; Stress--strain curve; Torque--speed curve; Pump characteristic curve; S--N fatigue curve.}

\label{fig:plotchain_montage}
\end{figure*}

% =========================
\section{Experimental Setup and Evaluation}
\label{sec:eval}

\subsection{Model Suite and Deterministic Decoding}
\label{sec:models}
We evaluate PlotChain using four state-of-practice multimodal LLMs spanning three providers:
(i) OpenAI GPT-4.1, (ii) OpenAI GPT-4o, (iii) Anthropic Claude Sonnet 4.5, and (iv) Google Gemini 2.5 Pro.
To maximize reproducibility, all evaluations use deterministic decoding (temperature $=0$) with a shared prompt template and a fixed maximum output token budget.
Run timestamps, model identifiers, and artifact checksums are recorded in the release manifest.

\begin{table}[t]
\centering
\caption{Evaluated models and run configuration (Jan.\ 2026; temperature $=0$ for all runs).}
\label{tab:models}
\setlength{\tabcolsep}{6pt}
\renewcommand{\arraystretch}{1.08}
\begin{tabular}{p{0.15\linewidth} p{0.24\linewidth} p{0.4\linewidth}}
\toprule
\textbf{Provider} & \textbf{Model} & \textbf{Notes} \\
\midrule
OpenAI & GPT-4.1 & Shared prompt; strict JSON-only numeric schema \\
OpenAI & GPT-4o & Same settings as above \\
Anthropic & Claude Sonnet 4.5 & Same settings; multimodal (image+text) input \\
Google & Gemini 2.5 Pro & Same settings; fixed high token budget in our implementation \\
\bottomrule
\end{tabular}
\end{table}

\subsection{Dataset-Driven Output Schema and Prompt}
\label{sec:prompt}
Each PlotChain item specifies an ordered set of expected output fields, partitioned into \emph{final} fields and \emph{checkpoint} fields.
Checkpoint fields (prefixed \texttt{cp\_}) are intermediate reads intended to diagnose where failures occur (e.g., reading a cutoff frequency correctly but computing a derived quantity incorrectly).
The evaluator constructs a strict JSON schema \emph{per item} from the dataset metadata and instructs the model to return \emph{only} a single JSON object containing numeric values (or \texttt{null}) for those keys.

\subsection{Robust Output Parsing}
\label{sec:parsing}
We enforce a strict interface (single JSON object), but we also implement robust extraction to ensure the benchmark measures plot-reading ability rather than brittle formatting errors.
Given the raw model response, the evaluator attempts: (i) direct JSON parsing, (ii) parsing a fenced \texttt{```json} block if present, and (iii) extracting the first \texttt{\{...\}} span.
To mitigate a common real-world failure mode where models output arithmetic expressions (e.g., \texttt{1025/615}) despite instructions, the parser sanitizes simple fractions into decimal values and removes trailing commas before re-parsing.
If parsing fails, all fields are treated as missing (equivalently \texttt{null}) for scoring.

\subsection{Tolerance-Based Numeric Scoring}
\label{sec:scoring}
Let $g$ denote the ground-truth numeric value for a field and $p$ the model-predicted value.
We compute absolute error $e_{\text{abs}} = |p-g|$ and relative error $e_{\text{rel}} = |p-g|/\max(|g|,\epsilon)$ with $\epsilon=10^{-12}$.
A prediction \emph{passes} if it falls within either an absolute tolerance $\tau_{\text{abs}}$ or a relative tolerance $\tau_{\text{rel}}$:
\begin{equation}
\label{eq:pass}
\textsc{Pass}(p,g) \;=\; \mathbb{I}\left[e_{\text{abs}} \le \tau_{\text{abs}} \;\;\vee\;\; e_{\text{rel}} \le \tau_{\text{rel}}\right].
\end{equation}

Tolerances are defined per (family, field) to reflect realistic human plot reading given axis resolution and visual affordances. We report all results under the \emph{plotread} tolerance policy.

\subsection{Metrics and Aggregation}
\label{sec:metrics}
We compute field-level, item-level, and model-level metrics, separating \emph{final} fields from \emph{checkpoint} fields:
\begin{itemize}
    \item \textbf{Field pass rate}: mean of Eq.~\eqref{eq:pass} over all items for a given (family, field).
    \item \textbf{Item final-pass}: an item passes if \emph{all} final fields pass.
    \item \textbf{Item checkpoint-pass}: an item passes if \emph{all} checkpoint fields pass.
    \item \textbf{Model final pass rate}: fraction of items that final-pass (headline metric).
    \item \textbf{Model checkpoint pass rate}: fraction of items that checkpoint-pass (diagnostic metric).
    \item \textbf{Error statistics}: mean absolute error and mean relative error computed over numeric fields (excluding missing values).
    \item \textbf{Latency}: per-call runtime as recorded by the evaluator.
\end{itemize}

In addition to aggregates, we persist the raw model outputs (\texttt{raw\_\{provider\}\_\{model\}.jsonl}) so that all metrics can be recomputed without additional API calls under alternative tolerance policies or parsing rules.

\subsection{Paired Significance Testing}
\label{sec:stats}
Because all models are evaluated on the \emph{same fixed set of items}, comparisons between models are paired.
For the headline \emph{binary} metric (item strict all-pass over final fields), we use McNemar's exact test on discordant item outcomes \cite{mcnemar1947} and apply Holm correction across all pairwise comparisons \cite{holm1979}.
We additionally report paired bootstrap 95\% confidence intervals for the difference in strict all-pass rates (resampling items with replacement) \cite{efron1993bootstrap}.
For the \emph{continuous} per-item metric (item final-field accuracy: mean of final-field pass indicators per item), we use paired $t$-tests and report Cohen's $d$ on paired differences \cite{cohen1988}, again with Holm correction across pairwise comparisons \cite{holm1979}.

% =========================
% Results
% =========================
\section{Results}
\label{sec:results}

We report PlotChain results under the \texttt{plotread} tolerance policy (Sec.~\ref{sec:scoring}) with deterministic decoding (temperature $=0$) and a strict numeric-JSON interface (Secs.~\ref{sec:prompt}--\ref{sec:parsing}).
. Each model is evaluated on the same fixed set of 450 items (15 families $\times$ 30 items). We report both:
(i) \emph{field-level pass rate} (fraction of individual numeric fields within tolerance), and
(ii) \emph{item-level strict all-pass} (an item passes only if \emph{all} final fields for that item pass).
Because each item typically contains multiple required numeric outputs, strict all-pass is a more discriminative (and more conservative) headline metric.

\begin{figure}[t]
\centering
\includegraphics[width=\columnwidth]{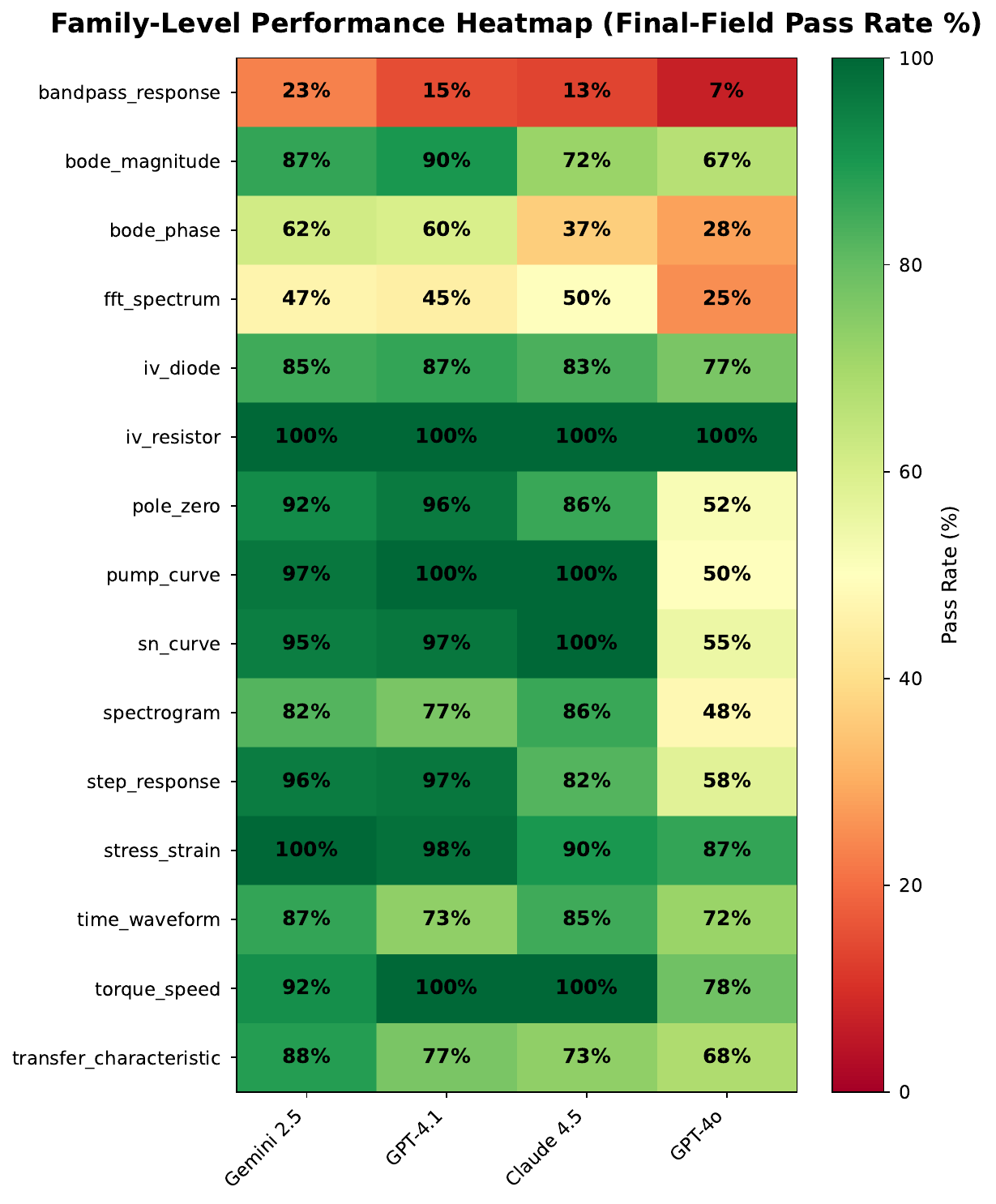}
\caption{Family-level performance heatmap (final-field pass rate, \%). Each cell is the percentage of \emph{final} (non-\texttt{cp\_*}) fields that pass tolerance, aggregated over all items in the family.}
\label{fig:family_heatmap}
\end{figure}

% -------------------------
% Overall performance
% -------------------------
\subsection{Overall performance and ranking}
\label{sec:overall_results}

Table~\ref{tab:overall_results} summarizes overall performance. On the headline item-level strict all-pass metric, Gemini~2.5~Pro ranks first (72.0\%), followed by GPT-4.1 (68.4\%), Claude Sonnet~4.5 (61.3\%), and GPT-4o (32.4\%).
Field-level pass rates are higher (as expected), with Gemini~2.5~Pro, GPT-4.1, and Claude Sonnet~4.5 clustered near $\sim$78--80\%, while GPT-4o is substantially lower.

\begin{table*}[t]
\centering
\caption{Overall PlotChain results under \texttt{plotread} tolerances (temperature $=0$). Field-level metrics are computed over all scored numeric fields. ``Final-field acc.'' is the per-item average accuracy over final (non-\texttt{cp\_*}) fields. ``Strict all-pass'' is the fraction of items where \emph{all final fields} pass.}
\label{tab:overall_results}
\renewcommand{\arraystretch}{1.08}
\setlength{\tabcolsep}{6.0pt}
\footnotesize
\begin{tabular}{lccccccc}
\toprule
\textbf{Model} &
\textbf{Field pass} &
\textbf{Final-field pass} &
\textbf{Checkpoint-field pass} &
\textbf{Final-field acc.} &
\textbf{Strict all-pass} &
\textbf{Checkpoint all-pass$^\dagger$} &
\textbf{Latency (s)} \\
\midrule
Gemini 2.5 Pro      & 80.4\% & 83.1\% & 76.5\% & 82.1\% & 72.0\% & 69.2\% & 20.66 \\
GPT-4.1             & 79.8\% & 81.9\% & 76.9\% & 80.7\% & 68.4\% & 70.8\% & 1.97 \\
Claude Sonnet 4.5   & 78.2\% & 77.7\% & 78.9\% & 77.1\% & 61.3\% & 75.4\% & 6.13 \\
GPT-4o              & 61.6\% & 57.0\% & 68.3\% & 58.0\% & 32.4\% & 59.2\% & 2.22 \\
\bottomrule
\end{tabular}

\vspace{1mm}
\footnotesize{$^\dagger$Checkpoint all-pass is computed over items that include at least one checkpoint field.}
\end{table*}

\begin{figure}[t]
\centering
\includegraphics[width=\columnwidth]{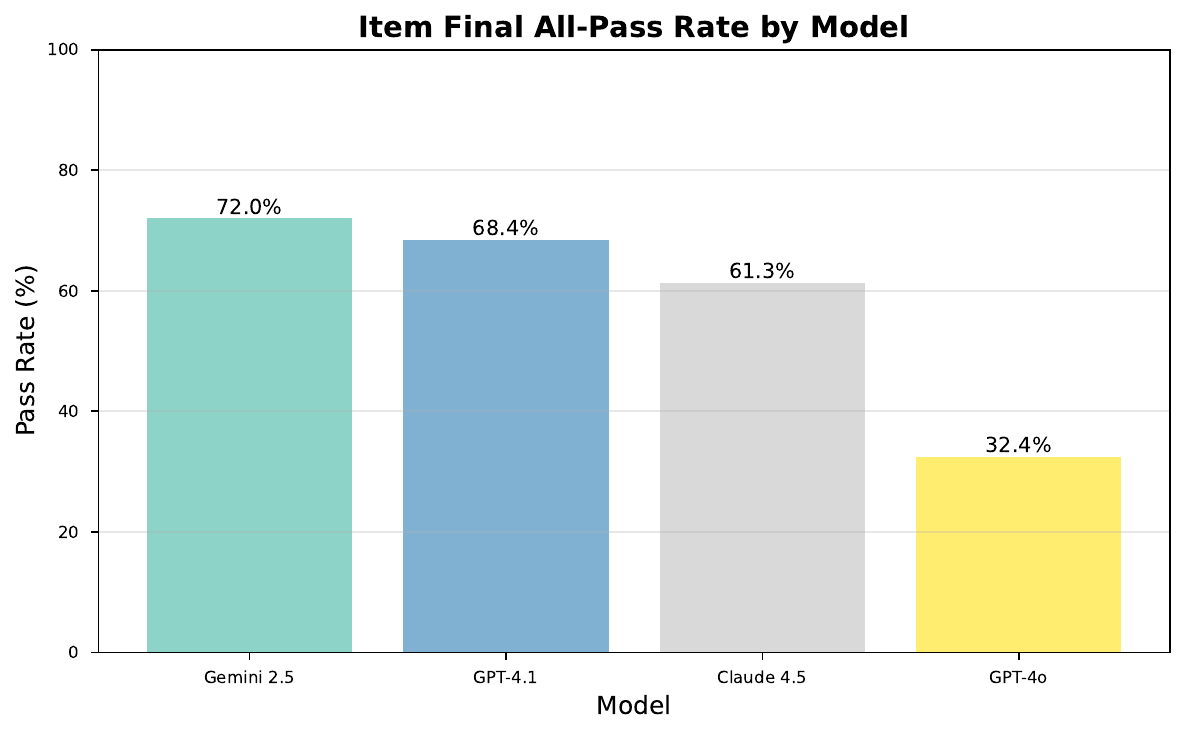}
\caption{Headline ranking by item-level strict all-pass (all final fields must pass).}
\label{fig:overall_allpass}
\end{figure}

% \begin{figure}[t]
% \centering
% \includegraphics[width=\columnwidth]{figures/accuracy_vs_latency.pdf}
% \caption{Accuracy--latency tradeoff using item-level strict all-pass vs.\ mean per-item latency.}
% \label{fig:acc_latency}
% \end{figure}

% -------------------------
% Family-level analysis
% -------------------------
\subsection{Family-level performance}
\label{sec:family_results}

Figure~\ref{fig:family_heatmap} reports \emph{family-level final-field pass rate} under \texttt{plotread}. Each cell aggregates pass/fail over the \emph{final} (non-\texttt{cp\_*}) numeric fields for all items in a plot family, providing a robust view of how often models recover the required end targets even when strict all-fields-per-item completion is not achieved.

Performance varies substantially across families. Several families are near-solved for frontier models, including \textsc{I--V (Resistor)} (100\% for all models) and high-performing response families such as \textsc{Bode Magnitude} (72--90\%). In contrast, frequency-domain and derived-quantity families remain the dominant bottleneck: \textsc{Bandpass Response} is consistently low (7--23\% across models) and \textsc{FFT Spectrum} remains challenging (25--50\%). These gaps indicate that quantitative extraction is not uniformly solved and that PlotChain retains diagnostic headroom across canonical engineering tasks.

% --- Wide figure: must be figure* in IEEE two-column ---
% \begin{figure*}[!t]
% \centering
% \vspace{-1.0ex}
% \includegraphics[width=\textwidth,height=0.28\textheight,keepaspectratio]{figures/family_heatmap.pdf}
% \vspace{-1.0ex}
% \caption{Per-family item-level strict all-pass (\%) under \texttt{plotread}. Darker cells indicate higher reliability on complete, multi-field plot-reading tasks within each family.}
% \label{fig:family_heatmap}
% \vspace{-1.0ex}
% \end{figure*}

% -------------------------
% Checkpoint diagnostics
% -------------------------
\subsection{Checkpoint diagnostics vs.\ final fields}
\label{sec:checkpoint_results}

\begin{figure}[t]
\centering
\includegraphics[width=\columnwidth]{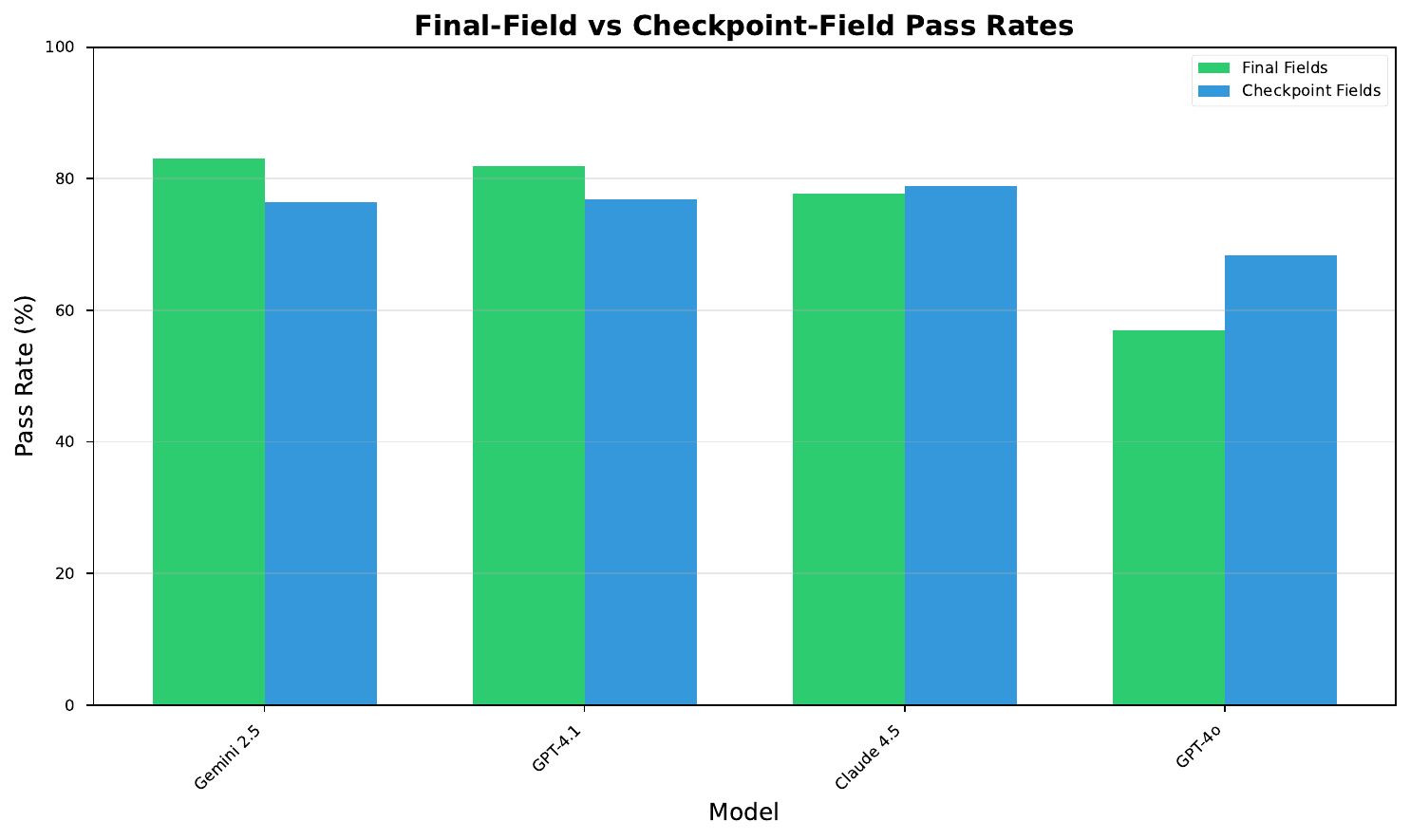}
\caption{Final vs.\ checkpoint field pass rates by model under \texttt{plotread}.}
\label{fig:final_vs_checkpoint}
\end{figure}

Checkpoint fields (\texttt{cp\_*}) are designed to localize failure causes by separating intermediate reads (e.g., crossing frequencies, peak locations) from final derived quantities. Figure~\ref{fig:final_vs_checkpoint} contrasts final vs.\ checkpoint field pass rates. Models differ systematically: GPT-4o and Claude Sonnet~4.5 tend to score relatively better on checkpoint fields than on final fields, consistent with extracting intermediate cues but failing to complete multi-step derivations reliably; Gemini~2.5~Pro and GPT-4.1 are comparatively stronger on final fields, suggesting fewer compounding errors from intermediate reads.

% -------------------------
% Paired statistics
% -------------------------
\subsection{Paired model comparisons}
\label{sec:paired_comparisons}

Because all models are evaluated on the same fixed item set, comparisons are paired. For the headline strict all-pass metric (binary per item), we use McNemar's exact test and report Holm-corrected $p$-values across all pairwise comparisons. Differences between the top three models are smaller than differences to GPT-4o; notably, Gemini~2.5~Pro vs.\ GPT-4.1 is not statistically significant at $\alpha=0.05$ after correction, whereas GPT-4.1 and Gemini~2.5~Pro both significantly outperform Claude Sonnet~4.5, and all three significantly outperform GPT-4o.

\begin{table}[t]
\centering
\caption{Paired comparisons on item-level strict all-pass (450 paired items). $\Delta$ is in percentage points (pp). McNemar uses exact binomial testing on discordant pairs, with Holm correction over all six pairwise tests.}
\label{tab:paired_allpass}
\renewcommand{\arraystretch}{1.08}
\setlength{\tabcolsep}{4pt}
\footnotesize
\begin{tabular}{p{0.4\linewidth} p{0.10\linewidth} p{0.2\linewidth} p{0.17\linewidth}}
\toprule
\textbf{Comparison} & $\boldsymbol{\Delta}$\textbf{ (pp)} & \textbf{95\% CI} & $\boldsymbol{p_\text{Holm}}$ \\
\midrule
Gemini 2.5 Pro vs GPT-4.1           & +3.6  & [ -0.2, +7.3 ] & 0.081 \\
GPT-4.1 vs Claude Sonnet 4.5        & +7.1  & [ +2.2, +11.8 ] & 0.0044 \\
Gemini 2.5 Pro vs Claude Sonnet 4.5 & +10.7 & [ +6.4, +14.9 ] & $3.8\times10^{-6}$ \\
Claude Sonnet 4.5 vs GPT-4o         & +28.9 & [ +24.0, +33.6 ] & $2.7\times10^{-27}$ \\
GPT-4.1 vs GPT-4o                   & +36.0 & [ +31.1, +40.9 ] & $3.1\times10^{-39}$ \\
Gemini 2.5 Pro vs GPT-4o            & +39.6 & [ +35.1, +44.2 ] & $3.4\times10^{-45}$ \\
\bottomrule
\end{tabular}
\end{table}

% ============================
% \section{Discussion: Failure Modes and Diagnostics}
% ============================
% =========================
% Discussion
% =========================
\section{Discussion}
\label{sec:discussion}

PlotChain is designed to diagnose \emph{plot-reading} rather than OCR-only transcription or generic captioning: models must infer quantitative values from axes, scales, and visual structure, and often combine multiple intermediate reads into derived quantities. The results show three main takeaways.

\subsection{Frontier models cluster on field-level accuracy, but diverge on strict end-to-end completion}
\label{sec:discussion_end_to_end}
At the field level, Gemini~2.5~Pro, GPT-4.1, and Claude Sonnet~4.5 cluster around $\sim$78--80\% pass rates (Table~\ref{tab:overall_results}), indicating that all three can often recover individual numeric targets from plots under human-realistic tolerances.
However, the item-level strict all-pass metric amplifies compounding errors: requiring \emph{all} final fields for an item to be correct reduces headline performance, with Gemini~2.5~Pro at 72.0\% and GPT-4.1 at 68.4\% (Table~\ref{tab:overall_results}).
Paired comparisons on strict all-pass confirm that the top-two gap is modest (Table~\ref{tab:paired_allpass}), while differences versus GPT-4o are substantial.
This illustrates why deterministic, multi-field items are valuable: they test reliable end-to-end completion rather than isolated ``some numbers right'' behavior.

\subsection{Bottlenecks concentrate in derived-quantity families}
\label{sec:discussion_bottlenecks}
Performance is highly family-dependent (Fig.~\ref{fig:family_heatmap}). Families that require reading multiple crossings or peaks and then computing a derived parameter remain challenging.
For example, \textsc{Bandpass Response} is unsolved under strict all-pass in our setting (0\% across models), even though models sometimes recover individual intermediate values within tolerance.
This pattern is consistent with a \emph{compounding-error} regime: small read errors in two or more dependent measurements (e.g., $f_1$ and $f_2$ at $-3$\,dB) can invalidate the derived bandwidth or $Q$ under strict all-pass.
In contrast, families with a single dominant readout or a clearer geometric structure (e.g., \textsc{I--V (Resistor)}) can be near-solved.

\subsection{Checkpoint fields localize failure modes beyond headline accuracy}
\label{sec:discussion_checkpoints}
Checkpoint fields (\texttt{cp\_*}) were introduced to distinguish failures in \emph{reading} the plot from failures in \emph{propagating} that read into downstream values.
Figure~\ref{fig:final_vs_checkpoint} shows that checkpoint vs.\ final field pass rates differ across models: some models extract intermediate cues more reliably than they finish multi-step derivations.
This enables actionable diagnostics: if a model passes checkpoints but fails finals, improvements should focus on arithmetic/derivation robustness and constraint following; if a model fails checkpoints, the bottleneck is upstream perception and axis calibration.
We view checkpoint-based scoring as complementary to conventional final-answer evaluation, and especially useful for tracking progress on specific failure sources as models evolve.

\subsection{Accuracy--latency tradeoffs matter for practical evaluation}
\label{sec:discussion_latency}
Our runs exhibit a clear accuracy--latency tradeoff (Table~\ref{tab:overall_results}). GPT-4.1 achieves competitive item-level strict all-pass with low latency (1.97\,s), while Gemini~2.5~Pro attains the highest strict all-pass at substantially higher latency (20.66\,s). Claude Sonnet~4.5 occupies an intermediate point (6.13\,s), and GPT-4o is fast but substantially less accurate (2.22\,s).
Such tradeoffs matter for practical engineering workflows where throughput and cost constraints can be as important as marginal gains in accuracy.

% =========================
% Limitations and Threats to Validity
% =========================
\section{Limitations and Threats to Validity}
\label{sec:limitations}

\subsection{Synthetic plots and coverage}
PlotChain uses a deterministic generator to produce ``gold'' ground truth from known parameters. This enables exact scoring and reproducibility, but synthetic plots may not capture all artifacts found in real lab/field plots (e.g., camera noise, compression, inconsistent styling, handwritten annotations, or overprinted legends).
We mitigate this by including multiple families, difficulty tiers, and edge-case renderings, but additional real-world evaluation remains future work.

\subsection{Tolerance dependence}
Our primary metric uses per-family/per-field tolerances intended to reflect realistic human plot reading. While this makes evaluation more faithful to the task, rankings can shift under different tolerance policies, especially for borderline reads near resolution limits.
To reduce ambiguity, we publish the tolerance policy, report both field- and strict item-level metrics, and release raw outputs to enable re-scoring without additional API calls.

\subsection{Interface constraints and compliance}
We require a strict numeric-JSON output schema to enable automated scoring. This is necessary for reproducibility, but it can penalize otherwise-correct reasoning if a model violates the interface (e.g., invalid JSON).
We partially address this via robust parsing and by retaining raw outputs for audit; nevertheless, strict all-pass reflects an end-to-end requirement that includes adherence to an evaluation contract.

\subsection{Model and platform variability}
Results reflect specific model versions and run configurations at the time of evaluation. Provider-side changes (model updates, serving stack changes, or safety filters) may affect outcomes.
We record run dates, settings, and checksums, and we release scripts and artifacts so results can be reproduced under the same conditions when possible. To preserve double-blind review, we will release the generator, evaluation scripts, tolerance policy, and raw outputs publicly upon acceptance.

\subsection{Scope of tasks}
PlotChain emphasizes numeric extraction and derived-quantity computation from canonical engineering plots. It does not directly test open-ended explanation quality, textual reasoning over captions, or tasks requiring domain-specific external knowledge beyond what is visible in the figure.
These are complementary capabilities and can be evaluated with other benchmarks alongside PlotChain.

% =========================
% Conclusion
% =========================
\section{Conclusion}
\label{sec:conclusion}

We introduced PlotChain, a deterministic, generator-based benchmark for evaluating multimodal LLMs on quantitative engineering plot reading. Unlike OCR-only or captioning-style evaluations, PlotChain pairs rendered plots with exact ground truth computed from generation parameters and augments final numeric targets with \texttt{cp\_*} checkpoint fields that expose intermediate reads for diagnostic analysis. The benchmark spans 15 canonical plot families (450 items total) with controlled difficulty and edge cases, enabling reproducible, fine-grained capability profiling across models.

Across four state-of-practice multimodal LLMs, we find that frontier models cluster on field-level pass rates, but diverge under strict end-to-end completion when multiple dependent outputs must be simultaneously correct. Family-level analyses reveal persistent bottlenecks in derived-quantity settings (e.g., frequency-response and spectrum-based tasks), while checkpoint metrics help distinguish upstream perception/axis-calibration failures from downstream derivation or error-compounding failures. Paired statistical comparisons confirm that the two top models in our study are statistically indistinguishable under \texttt{plotread} after multiple-comparison correction, while all three frontier models substantially outperform GPT-4o.

PlotChain and its evaluation artifacts are released to support transparent, repeatable benchmarking and to enable re-scoring under alternative tolerance policies without additional model queries. We hope PlotChain serves as a practical diagnostic tool for tracking progress in quantitative visual reasoning on engineering plots, and as a foundation for future extensions to broader plot styles, additional real-world noise sources, and new diagnostic families.

% ============================
\bibliographystyle{IEEEtran}
\bibliography{references}

% % ============================
% \appendices
% % ============================
% \section{Manual Readability Protocol (Summary)}
% \input{sections/A1_manual_readability.tex}

% \section{Prompt Template and JSON Schema}
% \input{sections/A2_prompt_schema.tex}

% \section{Additional Per-Family Results}
% \input{sections/A3_additional_results.tex}

\end{document}